\documentclass[11pt]{article}

\usepackage{acl}
\usepackage{times}
\usepackage{latexsym}
\usepackage[T1]{fontenc}
\usepackage[utf8]{inputenc}

\usepackage{microtype}

\usepackage{inconsolata}
\usepackage{graphicx}
\usepackage{amsmath} 
\usepackage{multirow}
\usepackage{subcaption}
\usepackage{makecell} 
\usepackage{fontspec}

\usepackage{listings}
\usepackage[T1]{fontenc}        
\usepackage{textcomp}           
\title{Retriv at BLP-2025 Task 1: A Transformer Ensemble and Multi-Task Learning Approach for Bangla Hate Speech Identification}
\author{
    Sourav Saha, K M Nafi Asib, and Mohammed Moshiul Hoque \\
    Department of Computer Science and Engineering \\
    Chittagong University of Engineering and Technology \\
    \{sahasourav1170, nafi.asib\}@gmail.com \\
    moshiul\_240@cuet.ac.bd
}

\begin{document}
\maketitle
\begin{abstract}

This paper addresses the problem of Bangla hate speech identification, a socially impactful yet linguistically challenging task. As part of the ``Bangla Multi-task Hate Speech Identification'' shared task at the BLP Workshop, IJCNLP--AACL 2025, our team ``Retriv'' participated in all three subtasks: (1A) hate type classification, (1B) target group identification, and (1C) joint detection of type, severity, and target. For subtasks 1A and 1B, we employed a soft-voting ensemble of transformer models (BanglaBERT, MuRIL, IndicBERTv2). For subtask 1C, we trained three multitask variants and aggregated their predictions through a weighted voting ensemble. Our systems achieved micro-$f_{1}$ scores of 72.75\% (1A) and 72.69\% (1B), and a weighted micro-$f_{1}$ score of 72.62\% (1C). On the shared task leaderboard, these corresponded to 9\textsuperscript{th}, 10\textsuperscript{th}, and 7\textsuperscript{th} positions, respectively. These results highlight the promise of transformer ensembles and weighted multitask frameworks for advancing Bangla hate speech detection in low-resource contexts. We made experimental scripts publicly available for the community.\footnote{\url{https://github.com/sahasourav17/Retriv-BLP25-Task-1}}
\end{abstract}

\section{Introduction}

With the rapid growth of social media platforms, harmful content such as hate speech and offensive language has become a pressing concern, requiring effective strategies to prevent its spread. Automated detection methods have seen substantial progress in high-resource languages, aided by large datasets and transformer-based models. However, in low-resource languages like Bangla hate speech detection remains challenging due to limited annotated resources, dialectal variation, and frequent code-mixing. Most existing work has focused on binary classification (hate vs.\ non-hate) or coarse multi-class labeling, leaving fine-grained dimensions such as \emph{type}, \emph{severity}, and \emph{target} underexplored. To address this gap, the BLP Workshop\footnote{\url{https://blp-workshop.github.io/}} at IJCNLP-AACL 2025~\cite{blp2025-overview-task1} introduced a shared task comprising three subtasks, including a multitask setup, to advance fine-grained hate speech modeling in Bangla. This paper advances current research by presenting our systems developed for the shared task. The key contributions of this work are illustrated in the following:

\begin{itemize}
    \item Proposed efficient yet competitive ensemble methods of Bangla-capable transformer models, achieving strong performance for fine-grained hate speech classification (Subtasks 1A and 1B).  
    \item Introduced a weighted voting ensemble within a multitask learning (MTL) framework, enabling joint prediction of hate \emph{type}, \emph{severity}, and \emph{target group}, and demonstrating the viability of MTL for Bangla hate speech.  
    \item Provided a comprehensive empirical study of deep learning and transformer-based approaches, including detailed performance comparisons and error analyses, offering insights for future research in low-resource hate speech detection.  
\end{itemize}

\section{Related Work}
 
Research on Bangla hate speech detection has expanded in recent years with several new datasets and modeling approaches. 
\citet{das-etal-2022-hate-speech} developed a corpus of Bangla and Romanized Bangla posts for hate and offensive language detection, demonstrating strong results with multilingual transformers such as XLM-R and MuRIL. 
\citet{saha2023vio} introduced Vio-lens, covering social media posts linked to communal violence, while \citet{haider-etal-2025-banth} proposed BanTH, a multi-label dataset for transliterated Bangla that captures multiple target categories and reflects the complexity of real-world hate speech. 
\citet{hasan2025llm} further introduced BanglaMultiHate, the first multi-task Bangla hate speech dataset jointly modeling \emph{type}, \emph{severity}, and \emph{target}, with extensive experiments using LLMs under zero-shot and LoRA fine-tuning. 
Beyond datasets, \citet{raza2024harmonynet} presented HarmonyNet, an ensemble framework that improves robustness in hate speech identification, and \citet{hossain2024align} proposed a multimodal approach aligning visual and textual features for hateful content detection. 
Despite these advances, most Bangla work remains focused on binary or coarse multi-class classification. 

Multi-task learning (MTL) has been explored as a way to capture complementary signals in hate speech detection. 
For example, \citet{awal2021angrybert} introduced AngryBERT, which jointly learns \emph{target} and \emph{emotion} alongside hate classification, while in the Bangla context, \citet{saha2024multsea} proposed MulTSeA, a multitask framework for aspect-based sentiment analysis. 
These studies suggest that MTL is well-suited for complex tasks like hate speech, where dimensions such as type, severity, and target are interdependent. 

Our work extends Bangla hate speech identification to a fine-grained, multi-dimensional setting. 
Unlike prior studies focused on binary or coarse classification, we employ transformer-based ensembles and a multitask framework to jointly model \emph{type}, \emph{severity}, and \emph{target group}, addressing a key gap in Bangla hate speech research. 

\section{Task and Dataset Descriptions}

The primary aim of this shared task~\cite{blp2025-overview-task1} is to conduct fine-grained hate speech identification in Bangla social media content, moving beyond binary detection toward multi-dimensional classification. The task was organized into multiple related subtasks:

\begin{itemize}
    \item \textbf{Subtask 1A (Hate Type)}: Classify a text as \emph{Abusive}, \emph{Sexism}, \emph{Religious Hate}, \emph{Political Hate}, \emph{Profane}, or \emph{None}.
    \item \textbf{Subtask 1B (Target Group)}: Identify whether the hate is targeted at \emph{Individuals}, \emph{Organizations}, \emph{Communities}, or \emph{Society}.
    \item \textbf{Subtask 1C (Multitask)}: Jointly predict the \emph{hate type}, \emph{severity} (\emph{Little to None}, \emph{Mild}, \emph{Severe}), and \emph{target group}.
\end{itemize}

All three subtasks share the same dataset splits with 35,522 samples for training, 2,512 for development, and 10,200 for testing. On average, the texts contain about 78 tokens across all splits. While the split sizes are identical, the label distributions differ across subtasks. Detailed label distributions for each subtask are provided in the Appendix~\ref{sec:dataset-analysis}

\section{System Description}

Our approach leverages different architectures tailored to the specific characteristics of each subtask. For subtasks 1A and 1B, we employ a soft ensemble of three pre-trained transformer models, while for subtask 1C, we adopt a  multitask learning framework. 

\subsection{Text Preprocessing}

We apply minimal preprocessing to preserve the authentic nature of social media content. The preprocessing pipeline consists of: (1) removal of Bangla digits, and (2) standard tokenization using each model's respective tokenizer. We observe that the provided dataset appears to be well-curated, requiring minimal additional cleaning steps. All input sequences are truncated or padded to a maximum length of 128 tokens.

\subsection{Base Models}

We utilize three complementary pre-trained transformer models across all subtasks:

\begin{itemize}
    \item \textbf{BanglaBERT} (\texttt{csebuetnlp/banglabert}): A monolingual BERT model specifically pre-trained on Bangla text, providing strong language-specific representations. \cite{bhattacharjee-etal-2022-banglabert}
    \item \textbf{MuRIL} (\texttt{google/muril-base-cased}): A multilingual model covering 17 Indian languages, including Bangla, offering cross-lingual contextual understanding. \cite{khanuja2021muril}
    \item \textbf{IndicBERTv2} (\texttt{ai4bharat/IndicBERTv2-MLM-only}): A model trained on 12 major Indian languages with enhanced tokenization for Indic scripts. \cite{doddapaneni-etal-2023-towards}
\end{itemize}

\subsection{Task-Specific Architectures}

\paragraph{Soft Voting Ensemble Approach}

For hate type classification (1A) and target group identification (1B), each base model is fine-tuned independently with task-specific classification heads. The final prediction is obtained through soft voting by averaging the prediction probabilities from all three models:

\begin{equation}
P_{\text{ensemble}}(y|x) = \frac{1}{3} \sum_{i=1}^{3} P_i(y|x)
\end{equation}

\noindent where $P_i(y|x)$ represents the probability distribution from model $i$ for input $x$.

\paragraph{Weighted Voting Multitask Approach}

For the joint prediction task, each base model is fine-tuned independently as a multitask learner with three classification heads: hate type (6 classes), severity (3 classes), and target group (4 classes). The individual multitask objective function combines cross-entropy losses from all tasks:

\begin{equation}
\mathcal{L}_{\text{mtl}} = \alpha \mathcal{L}_{\text{type}} + \beta \mathcal{L}_{\text{severity}} + \gamma \mathcal{L}_{\text{target}}
\end{equation}

The final ensemble prediction uses weighted voting based on individual development set performance:

\begin{equation}
\begin{aligned}
P_{\mathrm{final}}(y\mid x) &= 0.5\,P_{\mathrm{MuRIL}}(y\mid x) \\
&\quad+ 0.3\,P_{\mathrm{BanglaBERT}}(y\mid x) \\
&\quad+ 0.2\,P_{\mathrm{IndicBERTv2}}(y\mid x)
\end{aligned}
\end{equation}

\noindent where the weights reflect each model's individual performance ranking on the development set.

\subsection{Training Configuration}

All models are trained using the \verb|AdamW| optimizer with identical hyperparameters across all subtasks: learning rate of $2 \times 10^{-5}$, batch size of 16, and 3 training epochs.
For subtasks 1A and 1B, each model in the soft voting ensemble is trained independently with task-specific objectives. For subtask 1C, each model is trained independently as a multitask learner before combining predictions through weighted voting.

\section{Results and Analysis}

\subsection{Performance Against Baselines} 

Table~\ref{tab:results_all} reports system performance across all three subtasks, evaluated with the official metrics (micro-$f_{1}$ for 1A and 1B, and weighted micro-$f_{1}$ for 1C). The organizers also released three baselines-Random, Majority, and n-gram, which obtained 16.38, 56.38, and 60.20\% on 1A; 20.43, 59.74, and 62.09\% on 1B; and 23.04, 60.72, and 63.05\% on 1C, respectively.

\begin{table}[h]
\centering
\small
\begin{tabular}{l|cc|c}
\hline
\textbf{Model} & \multicolumn{2}{c|}{Micro-$f_{1}$} & W-Micro-$f_{1}$ \\
\cline{2-3}
                    & \textbf{1A} & \textbf{1B} & \textbf{1C} \\
\hline
BiLSTM (GV)         & 69.39 & 64.49 & -- \\
BiLSTM (FT)         & 68.67 & 62.74 & -- \\
BiGRU (GV)          & 69.35 & 68.75 & -- \\
BiGRU (FT)          & 66.92 & 68.75 & -- \\
MRL                 & 74.00 & 74.60 & 74.79 \\
BNB                 & 71.00 & 73.61 & 73.35 \\
INB                 & 74.00 & 73.17 & 71.22 \\
SV (MRL+BNB+INB)    & \textbf{75.72} & \textbf{74.96} & 74.08 \\
HV (MRL+BNB+INB)    & 72.53 & 74.16 & 73.42 \\
WV (MRL+BNB+INB)    & 74.16 & 74.56 & \textbf{75.12} \\
\hline
\end{tabular}
\caption{Performance of employed models across all subtasks. A dash (--) denotes that the model was not evaluated for the corresponding subtask. Abbreviations: GV=GloVe, FT=FastText, MRL=MuRIL, BNB=BanglaBERT, INB=IndicBERTv2, SV=Soft Voting, HV=Hard Voting, WV=Weighted Voting.}
\label{tab:results_all}
\end{table}

Our systems substantially outperform these baselines. For example, BanglaBERT attains 71.00\% on subtask 1A, more than 10 points higher than the n-gram baseline. The best-performing systems are ensembles: soft voting achieves 75.72\% (1A) and 74.96\% (1B), while weighted voting performs best on 1C (75.12\%).

\subsection{RNNs vs. Transformers vs. Ensembles}
On subtasks 1A and 1B, BiLSTM and BiGRU models with static embeddings (GloVe, FastText) yield scores in the mid-60s, significantly lower than transformer-based models. Based on these results, we did not extend RNNs to Subtask 1C. 

Among transformers, MuRIL and IndicBERTv2 perform comparably and generally surpass BanglaBERT, reflecting their larger multilingual training. However, ensemble methods consistently outperform individual models. Soft voting is most effective in 1A and 1B, whereas weighted voting excels in 1C, suggesting that the optimal ensemble strategy depends on task complexity.

\subsection{Error Analysis}
\paragraph{Confusion Patterns.} 
In subtask~1A (hate type), the system frequently confuses \textit{Abusive} with \textit{None}, and \textit{Political Hate} with \textit{Profane}. Minority categories such as \textit{Sexism} and \textit{Religious Hate} suffer from very low recall due to class imbalance. In subtask~1B (target group), \textit{Organization} is often predicted as \textit{None}, and \textit{Society} is confused with \textit{Individual}. Subtask~1C inherits these trends, with additional difficulty distinguishing between \textit{Mild} and \textit{Severe} hate severity. Confusion matrices and error examples are provided in Appendix~\ref{sec:detailed-error-analysis}.

\paragraph{Qualitative Errors.} 
Representative examples highlight typical misclassifications. For instance,  
\begin{itemize}
    \item \textbf{Implicit or sarcastic hate:} {\beng ভাইয়া আপনি অভিনেতা হইয়েন না না হলে সবাই বাচ্চা চাইবে} \textit{(Brother, don't act like an actor, otherwise everyone will demand children from you)} was annotated as \textit{Abusive, Individual}, but all systems predicted \textit{None}.  
    \item \textbf{Subtask complementarity:} {\beng এটা রে আওয়ামী লীগ ভোট দিবে কেনমাথায় আসে না} \textit{(Why would anyone vote for Awami League, I cannot imagine)} was misclassified in Subtask~1B (\textit{Organization}) but correctly resolved in the multitask model.  
    \item \textbf{Trade-offs in multitask learning:} {\beng হারামি মোসাদ্দেক দেখি এইখানে} \textit{(That bastard Mosaddek, I see him here)} was correctly identified as \textit{Profane, Individual} in single-task models, but misclassified as \textit{Abusive, Individual} in multitask.  
\end{itemize}

\paragraph{Effect of Label Imbalance.} 
The dataset distribution reveals severe class imbalance across subtasks (see Appendix~\ref{sec:dataset-analysis}). In Subtask~1A, categories such as \textit{Sexism} (only 122 training instances) and \textit{Religious Hate} (676 instances) are underrepresented, explaining their very low recall in our experiments. Similarly, in Subtask~1B, minority classes like \textit{Community} and \textit{Society} are frequently misclassified as \textit{None} or \textit{Individual}. For Subtask~1C, the dominance of the \textit{Little to None} category in severity prediction makes it challenging for models to correctly identify \textit{Mild} and \textit{Severe} hate. These imbalances highlight the need for data augmentation and re-weighting strategies in future work.

\subsection{Summary of Findings}
Our analysis shows that (i) transformer ensembles consistently outperform single models and RNN baselines, (ii) multitask learning captures complementary signals across hate type, severity, and target group, though sometimes introducing inconsistencies, and (iii) errors stem primarily from subtle linguistic cues, overlapping class boundaries, and severe class imbalance. Together, these findings validate the complementary strengths of ensemble and multitask strategies for Bangla hate speech identification.

\paragraph{Official Shared Task Results.}
On the blind test set used for leaderboard evaluation, our submissions achieved
72.75\% Micro-f1 in Subtask~1A (9th), 
72.69\% in Subtask~1B (10th), 
and 72.62\% Weighted Micro-f1 in Subtask~1C (7th).
These results confirm the competitiveness of our ensemble and multitask strategies in a challenging shared-task setting.

\section{Conclusion}

In this paper, we presented our systems for the BLP 2025 Shared Task 1 on Bangla hate speech identification. We explored transformer-based ensembles for subtasks 1A and 1B, and designed an efficient yet competitive multitask learning framework for subtask 1C. 
Our models achieved competitive performance across all subtasks, demonstrating the viability of both ensemble strategies and multitask learning in a low-resource setting like Bangla. 
The analysis further revealed challenges such as class imbalance and the difficulty of modeling underrepresented categories (e.g., \emph{Sexism}, \emph{Religious Hate}). 
For future work, we aim to explore data augmentation, cross-lingual transfer, and more robust multitask architectures to improve fine-grained hate speech detection in Bangla and extend these approaches to other low-resource languages.

\section{Limitations}

While our system demonstrates competitive performance on fine-grained hate speech detection, we acknowledge certain limitations. Our ensemble approaches require training and inference across multiple transformer models, increasing computational overhead compared to single-model solutions. The fixed weighting strategy for subtask 1C, while empirically determined from development performance, may benefit from more sophisticated dynamic weighting mechanisms. Additionally, our evaluation focuses on the shared task dataset, and broader cross-domain validation would strengthen the generalizability claims of our ensemble strategies. 

\bibliography{custom}

@article{hasan2025llm,
      title={LLM-Based Multi-Task Bangla Hate Speech Detection: Type, Severity, and Target}, 
      author={Hasan, Md Arid and Alam, Firoj and Hossain, Md Fahad and Naseem, Usman and Ahmed, Syed Ishtiaque},
      year={2025},
      journal={arXiv preprint arXiv:2510.01995},
      url={https://arxiv.org/abs/2510.01995},
}

@inproceedings{blp2025-overview-task1,
    title = "Overview of BLP 2025 Task 1: Bangla Hate Speech Identification",
    author = "Hasan, Md Arid and Alam, Firoj and Hossain, Md Fahad and Naseem, Usman and Ahmed, Syed Ishtiaque",
    booktitle = "Proceedings of the Second International Workshop on Bangla Language Processing (BLP-2025)",
    editor = {Alam, Firoj
          and Kar, Sudipta
          and Chowdhury, Shammur Absar
          and Hassan, Naeemul
          and Prince, Enamul Hoque
          and Tasnim, Mohiuddin
          and Rony, Md Rashad Al Hasan,
          and Rahman, Md Tahmid Rahman
    },
    month = dec,
    year = "2025",
    address = "India",
    publisher = "Association for Computational Linguistics",
}

@inproceedings{saha2024multsea,
  title={MulTSeA: Aspect-based Sentiment Analysis using Multitask Learning from Bengali Texts},
  author={Saha, Sourav and Ahsan, Shawly and Hoque, Mohammed Moshiul},
  booktitle={2024 27th International Conference on Computer and Information Technology (ICCIT)},
  pages={25--31},
  year={2024},
  organization={IEEE}
}

@inproceedings{haider-etal-2025-banth,
    title = "{B}an{TH}: A Multi-label Hate Speech Detection Dataset for Transliterated {B}angla",
    author = "Haider, Fabiha  and
      Shifat, Fariha Tanjim  and
      Ishmam, Md Farhan  and
      Sourove, Md Sakib Ul Rahman  and
      Barua, Deeparghya Dutta  and
      Fahim, Md  and
      Bhuiyan, Md Farhad Alam",
    editor = "Chiruzzo, Luis  and
      Ritter, Alan  and
      Wang, Lu",
    booktitle = "Findings of the Association for Computational Linguistics: NAACL 2025",
    month = apr,
    year = "2025",
    address = "Albuquerque, New Mexico",
    publisher = "Association for Computational Linguistics",
    url = "https://aclanthology.org/2025.findings-naacl.403/",
    doi = "10.18653/v1/2025.findings-naacl.403",
    pages = "7217--7236",
    ISBN = "979-8-89176-195-7"
}

@inproceedings{das-etal-2022-hate-speech,
    title = "Hate Speech and Offensive Language Detection in {B}engali",
    author = "Das, Mithun  and
      Banerjee, Somnath  and
      Saha, Punyajoy  and
      Mukherjee, Animesh",
    editor = "He, Yulan  and
      Ji, Heng  and
      Li, Sujian  and
      Liu, Yang  and
      Chang, Chua-Hui",
    booktitle = "Proceedings of the 2nd Conference of the Asia-Pacific Chapter of the Association for Computational Linguistics and the 12th International Joint Conference on Natural Language Processing (Volume 1: Long Papers)",
    month = nov,
    year = "2022",
    address = "Online only",
    publisher = "Association for Computational Linguistics",
    url = "https://aclanthology.org/2022.aacl-main.23/",
    doi = "10.18653/v1/2022.aacl-main.23",
    pages = "286--296"
}

@article{raza2024harmonynet,
  title={HarmonyNet: navigating hate speech detection},
  author={Raza, Shaina and Chatrath, Veronica},
  journal={Natural Language Processing Journal},
  volume={8},
  pages={100098},
  year={2024},
  publisher={Elsevier}
}

@inproceedings{awal2021angrybert,
  title={Angrybert: Joint learning target and emotion for hate speech detection},
  author={Awal, Md Rabiul and Cao, Rui and Lee, Roy Ka-Wei and Mitrovi{\'c}, Sandra},
  booktitle={Pacific-Asia conference on knowledge discovery and data mining},
  pages={701--713},
  year={2021},
  organization={Springer}
}

@inproceedings{saha2023vio,
  title={Vio-lens: A novel dataset of annotated social network posts leading to different forms of communal violence and its evaluation},
  author={Saha, Sourav and Junaed, Jahedul Alam and Saleki, Maryam and Sharma, Arnab Sen and Rifat, Mohammad Rashidujjaman and Rahouti, Mohamed and Ahmed, Syed Ishtiaque and Mohammed, Nabeel and Amin, Mohammad Ruhul},
  booktitle={Proceedings of the First Workshop on Bangla Language Processing (BLP-2023)},
  pages={72--84},
  year={2023}
}

@inproceedings{hossain2024align,
  title={Align before Attend: Aligning Visual and Textual Features for Multimodal Hateful Content Detection},
  author={Hossain, Eftekhar and Sharif, Omar and Hoque, Mohammed Moshiul and Preum, Sarah Masud},
  booktitle={Proceedings of the 18th Conference of the European Chapter of the Association for Computational Linguistics: Student Research Workshop},
  pages={162--174},
  year={2024}
}

@inproceedings{doddapaneni-etal-2023-towards,
    title = "Towards Leaving No {I}ndic Language Behind: Building Monolingual Corpora, Benchmark and Models for {I}ndic Languages",
    author = "Doddapaneni, Sumanth  and
      Aralikatte, Rahul  and
      Ramesh, Gowtham  and
      Goyal, Shreya  and
      Khapra, Mitesh M.  and
      Kunchukuttan, Anoop  and
      Kumar, Pratyush",
    editor = "Rogers, Anna  and
      Boyd-Graber, Jordan  and
      Okazaki, Naoaki",
    booktitle = "Proceedings of the 61st Annual Meeting of the Association for Computational Linguistics (Volume 1: Long Papers)",
    month = jul,
    year = "2023",
    address = "Toronto, Canada",
    publisher = "Association for Computational Linguistics",
    url = "https://aclanthology.org/2023.acl-long.693",
    doi = "10.18653/v1/2023.acl-long.693",
    pages = "12402--12426",
}

@inproceedings{bhattacharjee-etal-2022-banglabert,
    title = "{B}angla{BERT}: Language Model Pretraining and Benchmarks for Low-Resource Language Understanding Evaluation in {B}angla",
    author = "Bhattacharjee, Abhik  and
      Hasan, Tahmid  and
      Ahmad, Wasi  and
      Mubasshir, Kazi Samin  and
      Islam, Md Saiful  and
      Iqbal, Anindya  and
      Rahman, M. Sohel  and
      Shahriyar, Rifat",
    editor = "Carpuat, Marine  and
      de Marneffe, Marie-Catherine  and
      Meza Ruiz, Ivan Vladimir",
    booktitle = "Findings of the Association for Computational Linguistics: NAACL 2022",
    month = jul,
    year = "2022",
    address = "Seattle, United States",
    publisher = "Association for Computational Linguistics",
    url = "https://aclanthology.org/2022.findings-naacl.98/",
    doi = "10.18653/v1/2022.findings-naacl.98",
    pages = "1318--1327"
}

@article{khanuja2021muril,
  title={Muril: Multilingual representations for indian languages},
  author={Khanuja, Simran and Bansal, Diksha and Mehtani, Sarvesh and Khosla, Savya and Dey, Atreyee and Gopalan, Balaji and Margam, Dilip Kumar and Aggarwal, Pooja and Nagipogu, Rajiv Teja and Dave, Shachi and others},
  journal={arXiv preprint arXiv:2103.10730},
  year={2021}
}

\appendix

\section{Datasets Analysis}
\label{sec:dataset-analysis}

\subsection{Label-wise Distribution}
\label{subsection:label-wise-distribution}

\begin{table}[h!]
\centering
\small
\begin{tabular}{llrrr}
\hline
\textbf{Subtask} & \textbf{Label} & \textbf{Train} & \textbf{Dev} & \textbf{Test} \\
\hline
\multirow{6}{*}{\shortstack{1A \\ (Type)}} 
 & None            & 19,954 & 1,447 & 5,751 \\
 & Abusive         & 8,212  &   549 & 2,312 \\
 & Political Hate  & 4,227  &   283 & 1,220 \\
 & Profane         & 2,331  &   185 &   709 \\
 & Religious Hate  &   676  &    40 &   179 \\
 & Sexism          &   122  &     8 &    29 \\
\hline
\multirow{5}{*}{\shortstack{1B \\ (Target)}} 
 & None            & 21,190 & 1,528 & 6,093 \\
 & Individual      &  5,646 &   391 & 1,571 \\
 & Organization    &  3,846 &   292 & 1,152 \\
 & Community       &  2,635 &   159 &   759 \\
 & Society         &  2,205 &   142 &   625 \\
\hline
\multirow{6}{*}{\shortstack{1C \\ (Type)}} 
 & None            & 19,954 & 1,447 & 5,751 \\
 & Abusive         & 8,212  &   549 & 2,312 \\
 & Political Hate  & 4,227  &   283 & 1,220 \\
 & Profane         & 2,331  &   185 &   709 \\
 & Religious Hate  &   676  &    40 &   179 \\
 & Sexism          &   122  &     8 &    29 \\
\hline
\multirow{3}{*}{\shortstack{1C \\ (Severity)}} 
 & Little to None  & 23,489 & 1,714 & 6,737 \\
 & Mild            &  6,853 &   426 & 2,001 \\
 & Severe          &  5,180 &   372 & 1,462 \\
\hline
\multirow{5}{*}{\shortstack{1C \\ (Target)}} 
 & None            & 21,190 & 1,528 & 6,093 \\
 & Individual      &  5,646 &   391 & 1,571 \\
 & Organization    &  3,846 &   292 & 1,152 \\
 & Community       &  2,635 &   159 &   759 \\
 & Society         &  2,205 &   142 &   625 \\
\hline
\end{tabular}
\caption{Label distributions across Train, Dev, and Test splits for all subtasks.}
\label{tab:label-distribution}
\end{table}

Table~\ref{tab:label-distribution} presents the detailed label-wise distributions across train, development, and test sets for all subtasks. These statistics highlight the inherent class imbalance, such as the small number of \emph{Sexism} and \emph{Religious Hate} instances in Subtask~1A, which makes the task more challenging. 

\section{Detailed Error Analysis}
\label{sec:detailed-error-analysis}
We include confusion matrices for all three subtasks to illustrate misclassification patterns, along with representative failure cases that highlight challenges such as class imbalance, and subtle contextual cues.

\begin{figure}[th]
    \centering
    \includegraphics[width=0.97\linewidth]{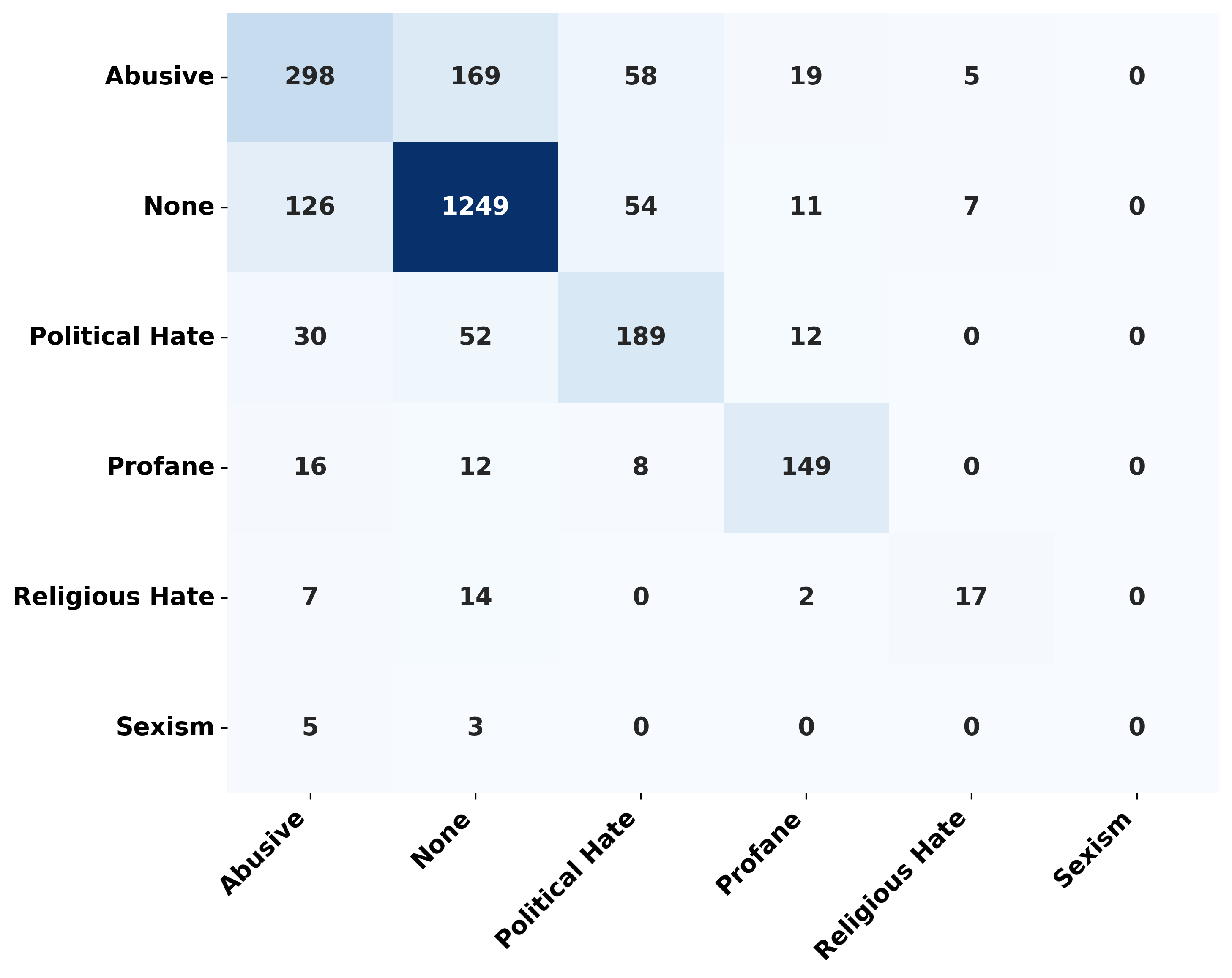}
    \caption{Confusion Matrix for Subtask 1A}
    \label{fig:1A_confusion_matrix}
\end{figure}

\begin{figure}[th]
    \centering
    \includegraphics[width=0.97\linewidth]{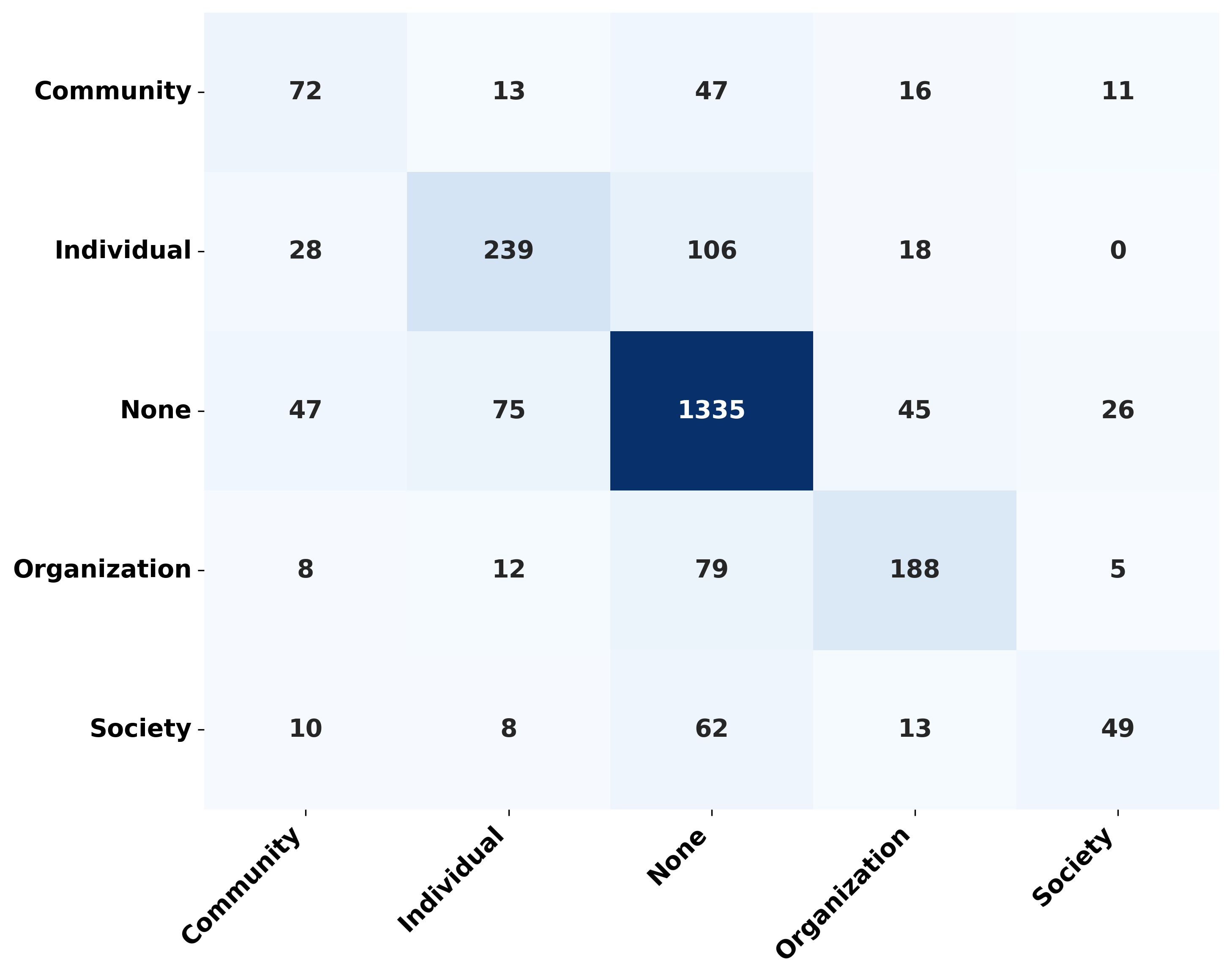}
    \caption{Confusion Matrix for Subtask 1B}
    \label{fig:1B_confusion_matrix}
\end{figure}


\begin{figure}[!thbp]
    \centering
    \begin{subfigure}{.48\textwidth}
      \centering
      \includegraphics[width=.98\linewidth]{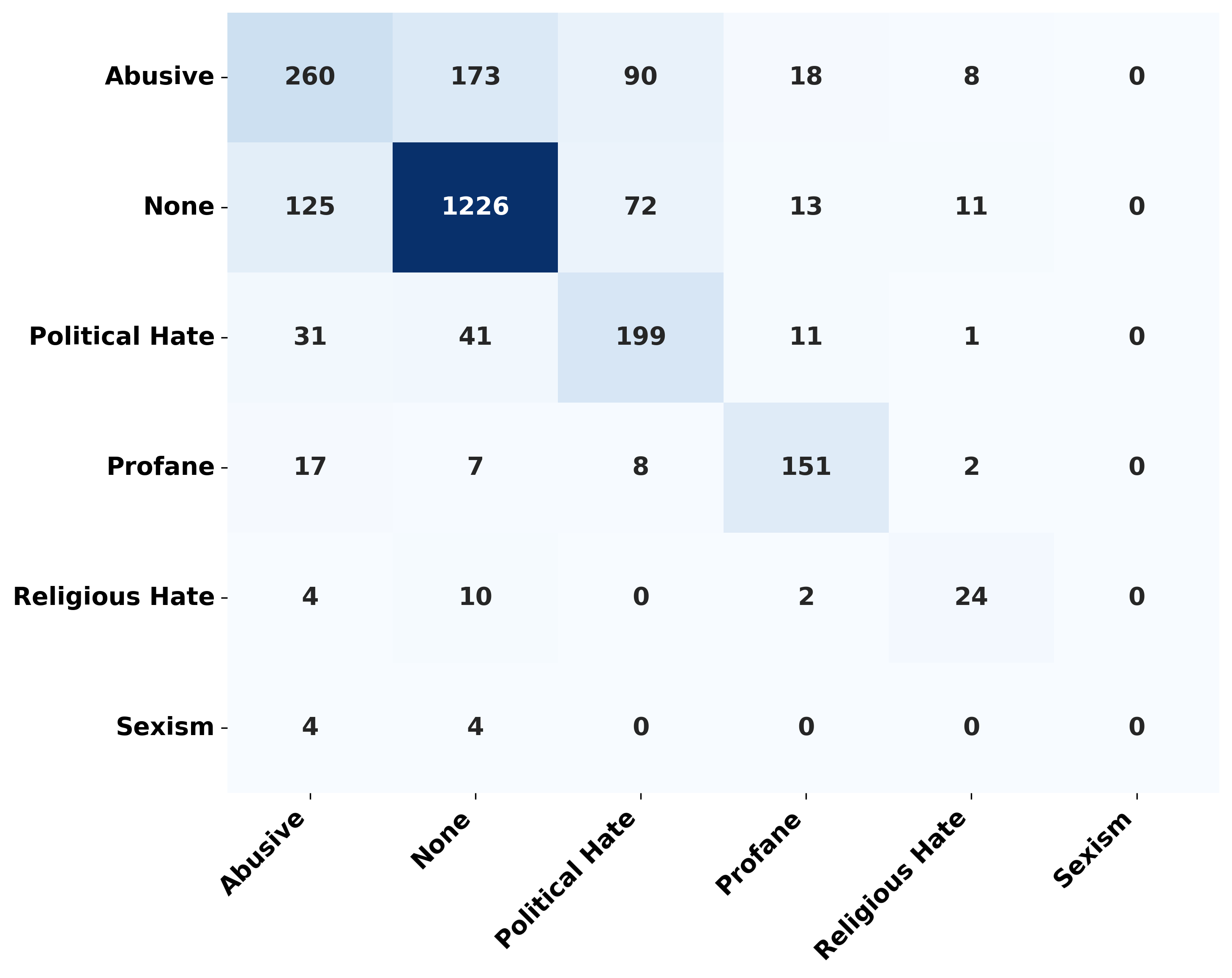}
      \caption{Hate Type}
      \label{fig:sub1}
    \end{subfigure}%
    \hfill
    \begin{subfigure}{.48\textwidth}
      \centering
      \includegraphics[width=.98\linewidth]{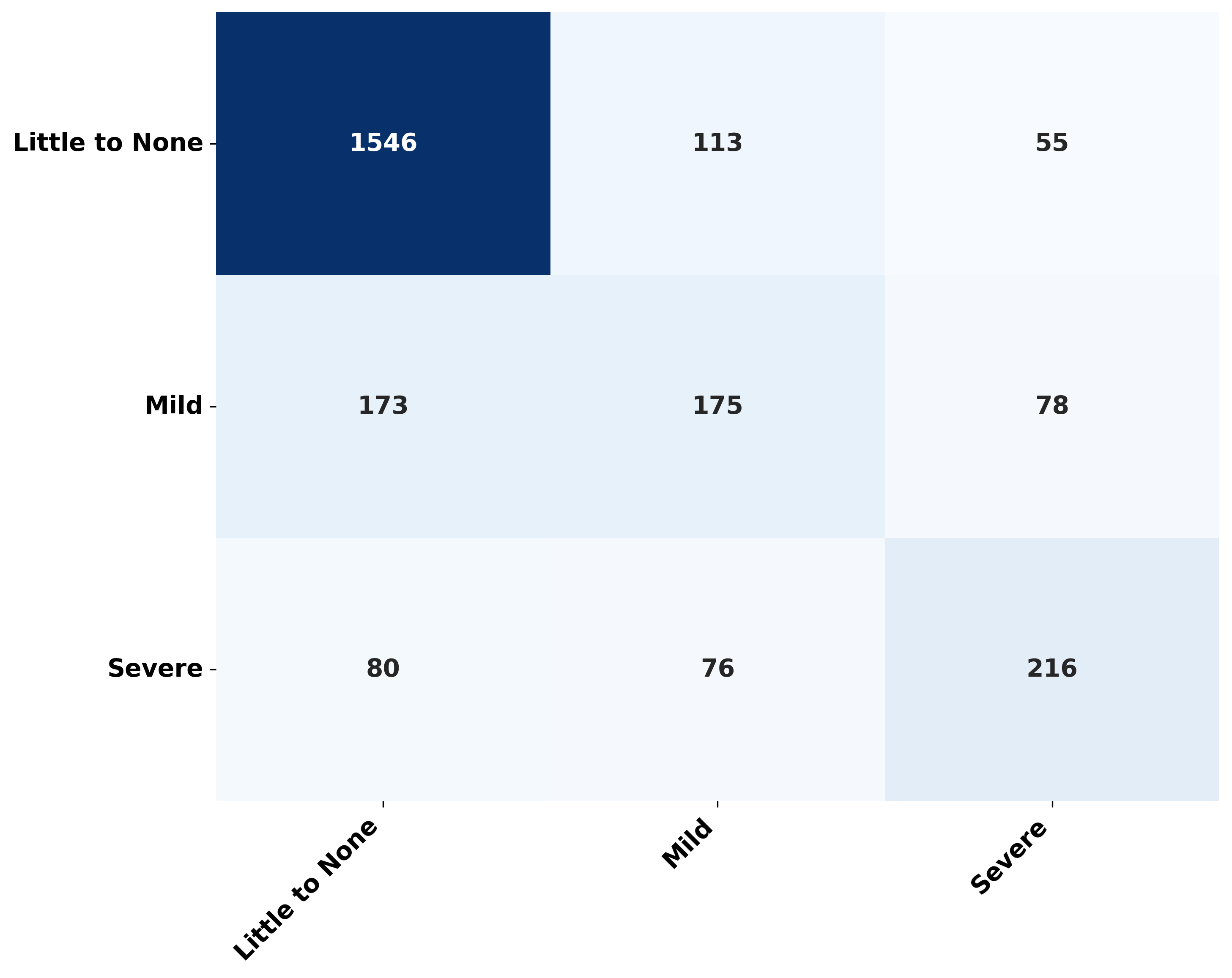}
      \caption{Hate Severity}
      \label{fig:sub2}
    \end{subfigure}%
    \hfill
    \begin{subfigure}{.48\textwidth}
      \centering
      \includegraphics[width=.98\linewidth]{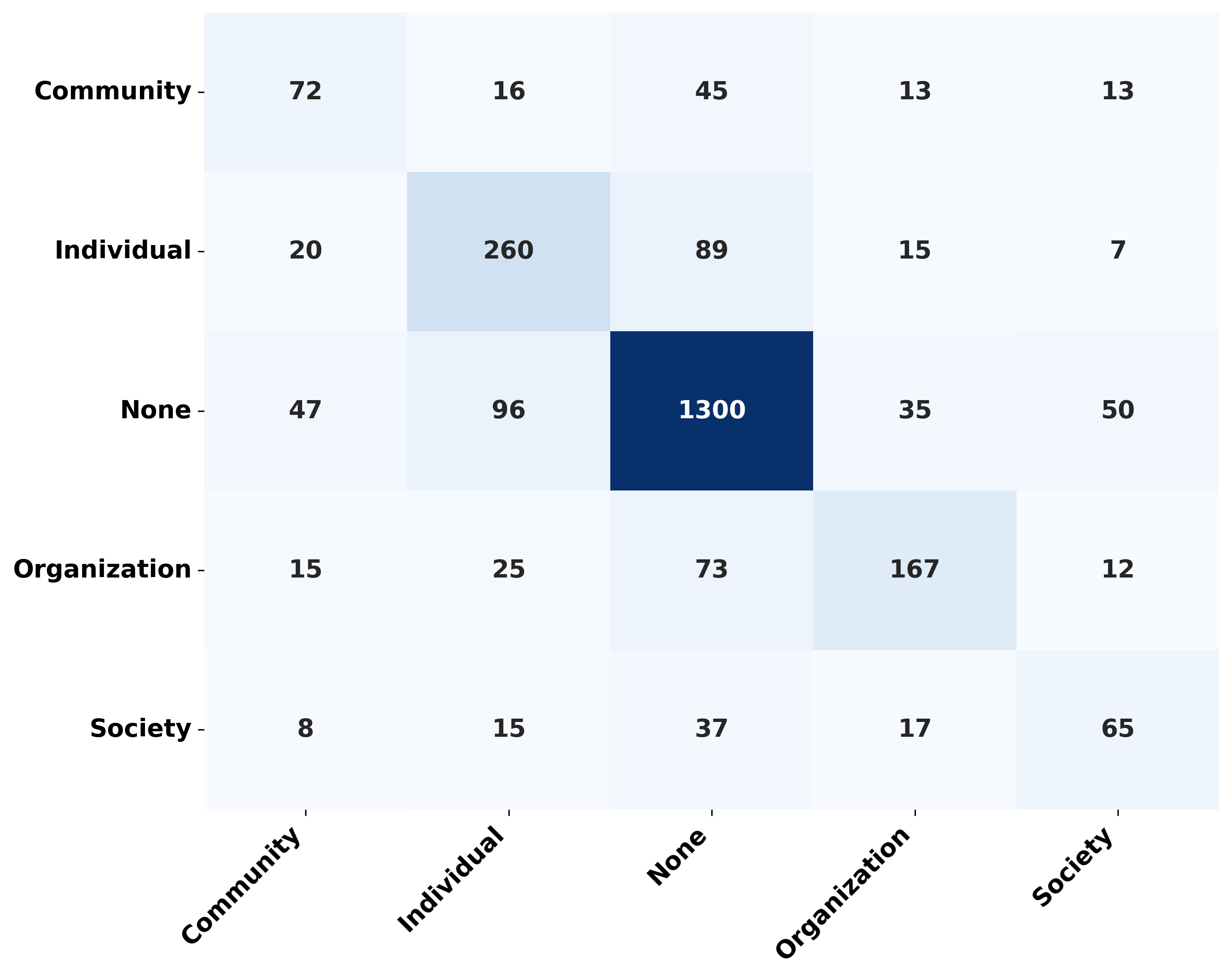}
      \caption{To Whom}
      \label{fig:sub3}
    \end{subfigure}
    
    \caption{Confusion matrices for subtask 1C}
    \label{fig:confusion_matrix}
\end{figure}

Since all three subtasks were derived from the same dataset, we analyze errors using a shared set of representative examples. 
This allows us to highlight systematic issues across subtasks in a more consistent manner. 
Instead of comparing systems directly against each other, we focus on cases where each subtask fails, succeeds, or faces systematic challenges. 
Wrong predictions are highlighted in blue. 

\subsection{Subtask 1A: Hate Type Classification}
Table~\ref{tab:errors-1a} shows cases where Subtask~1A (type classification) fails. 
We observe frequent confusion between \textit{Abusive} and \textit{Profane}, as well as under-prediction of subtle \textit{Political Hate}. 
These errors often arise from short texts or figurative language. 

\begin{table}[htbp]
\centering
\footnotesize
\resizebox{\linewidth}{!}{%
\begin{tabular}{p{0.45\linewidth} p{0.24\linewidth} p{0.24\linewidth}}
\hline
\textbf{Text} & \makecell{\textbf{Gold} \\ \textbf{Annotation}} & \makecell{\textbf{1A} \\ \textbf{Prediction}} \\
\hline
{\beng সয়তানর খালাম্মা কয়কি} \\
\textit{(What did the devil's aunt say?)} &
Abusive &
\textcolor{blue}{None} \\
\hline
{\beng নির্বাচক মণ্ডলীর দের কে খেলানো দরকার হালারা} \\
\textit{(The election committee members should be made to play, idiots.)} &
Political Hate &
\textcolor{blue}{Profane} \\
\hline
\end{tabular}
}
\caption{Examples where Subtask~1A (type) failed. Wrong predictions are in blue.}
\label{tab:errors-1a}
\end{table}

\subsection{Subtask 1B: Target Group Classification}
Table~\ref{tab:errors-1b} presents errors from Subtask~1B (target group identification). 
The main challenge lies in distinguishing \textit{Organization} vs. \textit{Community}, and in cases where hate is implied but the target is indirect. 

\begin{table}[htbp]
\centering
\footnotesize
\resizebox{\linewidth}{!}{%
\begin{tabular}{p{0.45\linewidth} p{0.24\linewidth} p{0.24\linewidth}}
\hline
\textbf{Text} & \makecell{\textbf{Gold} \\ \textbf{Annotation}} & \makecell{\textbf{1B} \\ \textbf{Prediction}} \\
\hline
{\beng তোমাদের রাজনীতি হাস্যকর} \\
\textit{(Your politics is ridiculous.}) &
Organization &
\textcolor{blue}{Community} \\
\hline
{\beng ভালো থাকলে কাপড় খুলে বের হস ... জাহান্নামে গেলি} \\
\textit{(If you're fine, strip off your clothes ... went to hell.)} &
Community &
\textcolor{blue}{Individual} \\
\hline
\end{tabular}
}
\caption{Examples where Subtask~1B (target) failed. Wrong predictions are in blue.}
\label{tab:errors-1b}
\end{table}

\subsection{Subtask 1C: Multitask Classification}
Table~\ref{tab:errors-1c} highlights errors in Subtask~1C (multitask). 
Although multitask modeling captures interdependencies between type, severity, and target, we observe systematic errors such as over-predicting \textit{Severe}, mismatched targets, and type drift. 

\begin{table}[htbp]
\centering
\footnotesize
\resizebox{\linewidth}{!}{%
\begin{tabular}{p{0.45\linewidth} p{0.24\linewidth} p{0.24\linewidth}}
\hline
\textbf{Text} & \makecell{\textbf{Gold} \\ \textbf{Annotation}} & \makecell{\textbf{1C} \\ \textbf{Prediction}} \\
\hline
{\beng গোয়া মারা দিয়ে আছে বাংলাদেশ মাদারচোদ নিউজ করে সালার পুত পাম দেস} \\
\textit{(Bangladesh is ruined motherfucker making news and buttering them up.)} &
Profane / Mild / Organization &
Profane / \textcolor{blue}{Severe} / \textcolor{blue}{Individual} \\
\hline
{\beng হারামি মোসাদ্দেক দেখি এইখানে} \\
\textit{(That bastard Mosaddek, I see him here.)} &
Profane / Little to None / Individual &
\textcolor{blue}{Abusive} / \textcolor{blue}{Severe} / Individual \\
\hline
\end{tabular}
}
\caption{Examples where Subtask~1C (multitask) failed. Wrong predictions are in blue.}
\label{tab:errors-1c}
\end{table}

\subsection{Cases Where All Subtasks Fail}
Finally, Table~\ref{tab:errors-all} shows difficult examples where all systems fail. 
These include sarcasm, implicit hate, or ambiguous targets, which remain challenging for current transformer models. 

\begin{table}[!htbp]
\centering
\footnotesize
\resizebox{\linewidth}{!}{%
\begin{tabular}{p{0.45\linewidth} p{0.24\linewidth} p{0.24\linewidth}}
\hline
\textbf{Text} & \makecell{\textbf{Gold} \\ \textbf{Annotation}} & \makecell{\textbf{System} \\ \textbf{Predictions}} \\
\hline
{\beng দনের নিবাচন ফাইজলামি৷} \\
\textit{(Today's election is a farce.)} &
Profane / Mild / None &
\textcolor{blue}{None / None / None} \\
\hline
{\beng তোমরা শুধু প্রতিবাদে জানাতে পারবে ... জনগণের টাকা দিয়ে খুজতেছে জনগণকে রক্ষা করার জন্য} \\
\textit{(You will only be able to protest ... wasting people's money pretending to protect them.)} &
Abusive / Little to None / Community &
\textcolor{blue}{Political Hate / None / None} \\
\hline
\end{tabular}
}
\caption{Challenging examples where all subtasks failed. Wrong predictions are in blue.}
\label{tab:errors-all}
\end{table}

\section{Additional Training Details}
\label{sec:training-appendix}
The training configuration used in our experiments is summarized in Table~\ref{tab:training-config}.

\begin{table}[!h]
\centering
\begin{tabular}{l c}
\hline
\textbf{Parameter} & \textbf{Value} \\
\hline
Optimizer          & AdamW \\
Learning rate      & $2 \times 10^{-5}$ \\
Batch size         & 16 \\
Epochs             & 3 \\
Max sequence length& 128 \\
\hline
\end{tabular}
\caption{Training hyperparameters used across all transformer models.}
\label{tab:training-config}
\end{table}

\section{Reproducibility Note}

All experiments were conducted on a single NVIDIA RTX~3090 GPU with 24\,GB VRAM. 
We used the PyTorch\footnote{\url{https://pytorch.org/}} deep learning framework together with the HuggingFace\footnote{\url{https://huggingface.co/}} Transformers library. 
All hyperparameters are listed in Table~\ref{tab:training-config}. 
Random seeds were fixed across runs for consistency, although minor variations in results may occur due to non-deterministic GPU operations.

\end{document}